\documentclass{article}

\usepackage{arxiv}

\usepackage[utf8]{inputenc} 
\usepackage[T1]{fontenc}    
\usepackage[hidelinks]{hyperref}
\usepackage{url}            
\usepackage{booktabs}       
\usepackage{amssymb, amsmath}
\usepackage{amsfonts}       
\usepackage{nicefrac}       
\usepackage{microtype}      
\usepackage{xfrac}
\usepackage{microtype}
\usepackage{graphicx}
\usepackage{subfigure}
\usepackage{bm}
\usepackage[table]{xcolor}
\usepackage{color}
\usepackage{pdfpages}
\usepackage{float}
\usepackage{natbib}
\usepackage{algorithm}
\usepackage{algorithmic}
\usepackage{framed}
\usepackage{enumitem}
\usepackage{cancel}
\usepackage{changes}
\usepackage[export]{adjustbox}
\setcitestyle{authoryear,open={(},close={)}}

\newtheorem{proposition}{Proposition}

\def\Sb{{\bm S}}

\def\Wb{{\bm W}}

\def\0{{\mathbf 0}}

\newcommand{\Ccal}{\mathcal{C}}

\newcommand{\Ncal}{\mathcal{N}}
\newcommand{\Mcal}{\mathcal{M}}
\newcommand{\Dcal}{\mathcal{D}}

\newcommand{\Rcal}{\mathcal{R}}
\newcommand{\Scal}{\mathcal{S}}

\newcommand{\Zcal}{\mathcal{Z}}

\newcommand{\xc}{\bm{x}}

\newcommand{\zc}{\bm{z}}

\newcommand{\maxf}[1]{{\cellcolor[gray]{0.87}} #1}

\title{On Masked Pre-training and the Marginal Likelihood}

\author{%
  Pablo Moreno-Mu\~{n}oz ~~~~~ Pol G. Recasens\thanks{Visiting student from Universitat Polit{e}cnica de Catalunya (\textsc{upc}), Barcelona.} ~~~~~~ Søren Hauberg\\ 
  Section for Cognitive Systems \\
  Technical University of Denmark (DTU)\\
  \texttt{\{pabmo,pgare,sohau\}@dtu.dk} \\
}

\begin{document}

\maketitle

\begin{abstract}
Masked pre-training removes random input dimensions and learns a model that can predict the missing values. Empirical results indicate that this intuitive form of self-supervised learning yields models that generalize very well to new domains. A theoretical understanding is, however, lacking. This paper shows that masked pre-training with a suitable cumulative scoring function corresponds to maximizing the model's marginal likelihood, which is \emph{de facto} the Bayesian model selection measure of generalization. Beyond shedding light on the success of masked pre-training, this insight also suggests that Bayesian models can be trained with appropriately designed self-supervision. Empirically, we confirm the developed theory and explore the main learning principles of masked pre-training in large language models.\looseness=-1
\end{abstract}

\section{Introduction}
Masked pre-training (\textsc{mpt}) is a family of self-supervised learning methods \citep{dosovitskiy2020image,devlin2018bert,caron2021emerging}, that empirically has been demonstrated to result in models that generalize very well to new settings. In essence, masked pre-training removes random features of the data and learns a model to recover these from the remaining input. While empirical results are impressive, a deeper understanding of \emph{why} pre-trained models generalize so well is lacking. Is it due to the use of transformer architectures \citep{vaswani2017attention}, the vast over-parametrization \citep{srebro2019role}, or something entirely different? 

The marginal likelihood or \emph{evidence} is commonly used as the measure of generalization ability in Bayesian models \citep{tenenbaum2001generalization,mackay2003information}. While computationally expensive, the \emph{blessing} of the marginal likelihood comes from the probabilistic integration of hypotheses. Whenever we are considering a latent variable model in the Bayesian framework, such integration can be thought of as the average over all the possible latent variable mappings, weighted by our prior beliefs. Since masked pre-training drives generalization so well, the lingering question in the Bayesian modeling community is then: \emph{Is masked pre-training somehow related to the maximization of the marginal likelihood?} 

\textbf{In this paper}, we provide a positive answer. We show that masked pre-training optimizes according to a stochastic gradient of the log-marginal likelihood (\textsc{lml}). Importantly, the log-marginal likelihood is equivalent to the cumulative sum of masked pre-training losses shaped with different sizes for the random mask. Even if its practical use avoids this cumulative sum, we show that choosing a \emph{fixed} masking rate, e.g.\@ $15\%$ as in \textsc{bert} \citep{devlin2018bert}, leads to a stochastic \emph{biased} estimation which still maximizes the log-marginal likelihood.

\textbf{Our proof} relies on a previous observation from \citet{fong2020marginal}, who shows that the log-marginal likelihood equals the average of exhaustive \emph{leave-$M$-out} cross-validation (\textsc{cv}) given posterior predictive scores. Intuitively, our formal results can be seen as the \emph{transposed} version of \citeauthor{fong2020marginal}'s results: where \textsc{cv} removes \emph{random observations} to measure generalization, masked pre-training removes \emph{random features}. While the seminal link between \textsc{cv} and the marginal likelihood was purely a formal result that pointed out the underlying presence of Bayesian principles in a well-known class of learning, our work extends the theory behind the marginal likelihood to comprehend the impressive behavior of the latest generative models.

\section{Masked pre-training}
\label{sec:mpt}

Masked pre-training (\textsc{mpt}) is a variant of self-supervised learning \citep{dosovitskiy2020image, devlin2018bert} that removes random input dimensions (also known as \emph{masking}) in the observed data and learns a model that accurately predicts the missing values. This family of methods, well-known due to their success in natural language understanding, typically adopts a transformer architecture \citep{vaswani2017attention} as the feature extractor, that together with positional encodings and random masked dimensions allows capturing the bidirectional context in the data.

In \textsc{bert} \citep{devlin2018bert}, each sentence is usually considered as a $D$ dimensional observation vector, $\xc = (x_1, x_2, \dots, x_D)^{\top}$, where dimensions $x_t$ are named \emph{tokens}. Given a \emph{random mask} $\mathcal{M}$ of size $M$$<$$D$, as a set of indices drawn uniformly from $\{1, \ldots, D\}$, each token whose index belongs to $\Mcal$ is considered to be in the subset $\xc_\Mcal = \{x_{\Mcal(1)}, x_{\Mcal(2)}, \dots, x_{\Mcal(M)}\}$. We refer to these as the \emph{masked tokens}. The rest of indices $\mathcal{R} = \{1,2,\dots, D\}\setminus \Mcal$ induce the complementary subset $\xc_{\Rcal}$, such that $\xc = \xc_\Mcal \cup \xc_{\Rcal}$. Under this notation, \textsc{mpt} learns the parameters $\theta$ of a model $p_{\theta}(\cdot)$ by maximising an average of the following objective
\begin{equation}
\log p_{\theta}(\xc_\Mcal|\xc_{\Rcal}) = \sum^{M}_{t=1}\log p_{\theta}(x_{\Mcal(t)}|\xc_{\Rcal})
\label{eq:mpt_loss}
\end{equation}
for every observation in the dataset $\Dcal$. The stochastic choice of $\xc_\Mcal$ makes predictive conditionals $p_{\theta}(\xc_\Mcal|\xc_{\Rcal})$ to be different for every observation and training step. Once the pre-training of $\theta$ has converged, this naturally allows the model to capture the underlying structure between dimensions of the data. One additional remark is the number of random masks needed to cover all combinations between \emph{masked} and observed tokens, which can be obtained as $\mathcal{C}_{M} = \binom{D}{M}$. In the particular example of \textsc{bert}, where the masking rate is $15\%$ with $D=512$ and $M=76$, the total number of random masks needed to cover all combinations of tokens is $\mathcal{C}_{M} \approx 1.21$$\times$$10^{92}$. This shows the inner combinatorial problem behind \textsc{mpt}. We provide empirical results on why \underline{this is not a limitation} for learning with \textsc{mpt} in Sec.~3.1.

\section{A probabilistic perspective, theory, and analysis}

Our key objective is to demonstrate that the good generalization of \textsc{mpt} can be explained as an equivalence with the model's high marginal likelihood. Indeed, we will prove that \textsc{mpt} \emph{implicitly} maximizes marginal likelihood according to some latent variable model of the form $p_\theta(\xc|\zc)$.

\paragraph{Marginal likelihood.} For our theory, we consider some dataset $\Dcal$ consisting of $n$ i.i.d.\@ observations $\xc_{1:n}$, where each sample $\xc_i$ could be either continuous or discrete and is of dimensionality $D$. We also assume that there exists a latent space $\Zcal \in \mathbb{R}^{K}$ where we can find unobserved variables $\zc_{1:n}$ which are part of the generative process of the data. This assumption is inspired in the common use of latent encodings in recent models fitted with \textsc{mpt}. In this direction, we also consider the observations to be samples of a likelihood function $p_{\theta}(\xc|\zc)$, where the mapping between the latent and observed variable is controlled by some parameters $\theta$, which might also include likelihood or prior \emph{hyperparameters}.

Importantly, we consider the parameters $\theta$ to be \emph{deterministic}, while we are interested in integrating out the latent variables that we cannot observe. Automatically, this leads us to the log-marginal likelihood (\textsc{lml}) of the model, which may factorize as a sum of marginals and can be also written as $\log p_\theta(\xc_{1:n}) = \sum^{n}_{i=1}\log p_\theta(\xc_{i})$, where the $i^{\text{th}}$ probability density comes from the integral $p_\theta(\xc_{i}) = \int p_{\theta}(\xc_{i}|\zc_{i})p(\zc_{i})\mathrm{d}\zc_{i}$. This definition coincides with the target \textsc{lml} used in the lower bound of variational autoencoders (\textsc{vae}) \citep{kingma2013auto,rezende2014stochastic} and it is widely used in probabilistic generative models.

\paragraph{Masking and conditional probabilities.} From the properties of probability distributions, we can decompose the individual \textsc{lml} functions $\log p_\theta(\xc_{i})$ as a sum of log-conditionals between \emph{tokens}. Omitting the $i^{\text{th}}$ observation subscript in $\xc$ to keep the notation uncluttered, the sum takes the form
\begin{equation}
\log p_{\theta}(\xc) = \sum^{D}_{t=1}\log p_{\theta}\left(x_t|\xc_{t+1:D}\right).
\label{eq:masked_tokens}
\end{equation}
However, the previous sum imposes a particular order on the selection of variables for conditioning, e.g.\@ $\{x_1|x_2, x_3, \dots\}, \{x_2|x_3, x_4, \dots\}$, etc. Moreover, the order of tokens in the observation vector remains predetermined, as dimensions are \underline{not} \emph{exchangeable}. Thus, we can consider a different combination of conditional probabilities in the sum --- for instance, $\{x_4|x_1, x_2, \dots\}, \{x_3|x_1, x_2, \dots\}$, etc. Here, the key insight is that the rules of probability applied to the log-marginal likelihood make it \emph{invariant} to the combination of different conditional factors, as we are observing different views of the same graphical model.

This combinatorial process between tokens in $\xc$ can be understood as the selection problem of indices. For that reason, we can assume a mask $\Mcal$ of the largest size $|\Mcal|=D$, such that $\Mcal \equiv \{1,2,\cdots, D\}$. Using similar properties of combinatorics, we can also obtain $D!$ different choices for $\Mcal$. While \emph{all} the indices are always in the set, the order of indices differs between combinations. This principled order in $\Mcal$ indicates how we sum the conditional probabilities in Eq.~\ref{eq:masked_tokens}. 

Since the \textsc{lml} is invariant to random choices of $\Mcal$, we can re-write the sum in Eq.~\ref{eq:masked_tokens} as an expectation with a \emph{countable set} of possible outcomes. Each outcome corresponds to one of the $D!$ choices for $\Mcal$, such that\looseness=-1
\begin{equation}
\log p_{\theta}(\xc) = \frac{1}{D!}\sum_{\pi=1}^{D!}\sum^{D}_{t=1}\log p_{\theta}\left(x^{(\pi)}_{\Mcal(t)}|\xc^{(\pi)}_{\Mcal(t+1:D)}\right) = \sum^{D}_{t=1} \mathbb{E}_{\pi}\left[\log p_{\theta}(x^{(\pi)}_{\Mcal(t)}|\xc^{(\pi)}_{\Mcal(t+1:D)})\right],
\label{eq:masked_tokens_v2}
\end{equation}
where the superscript $(\pi)$ denotes which mask $\Mcal$ are we using for indexing the tokens. We also \emph{swapped} the order of the sums to obtain the desired expectation in the r.h.s.\@ of the formula. 

\paragraph{The role of random masking.}
If we now take a particular index $(t)$ and we look at the $\pi^{\text{th}}$ summand in the previous expression, we can see that the \textsc{lml} is still \emph{invariant} to the order of the conditioning tokens $\xc_{\Mcal(t+1:D)}$ in the log-probabilities: $\log p_{\theta}(x_{\Mcal(t)}|\xc_{\Mcal(t+1:D)})$ in the sum. Intuitively, we can use both --- $\{x_1|x_2\},\{x_2\}$ or $\{x_2|x_1\},\{x_1\}$; independently of the conditional factors previously considered. In practice, this indicates that we can insert a \emph{second set} of indices to the r.h.s.\@ variables, which is the key point to link negative \textsc{mpt} loss and \textsc{lml}.

Now, assume that $\Mcal$ indexes less than $100\%$ of tokens, while the rest is indexed by $\mathcal{R}$ as defined in Sec.~\ref{sec:mpt}. If we match both complementary masks to be aligned with the \emph{conditional} and \emph{conditioning} variables in the log-probabilities, this allows us to rewrite the $t^{\text{th}}$ summands in Eq.~\ref{eq:masked_tokens} as 
$$\frac{1}{D!}\sum_{\pi=1}^{D!}\log p_{\theta}\left(x^{(\pi)}_{\Mcal(t)}|\xc^{(\pi)}_{\Mcal(t+1:D)}\right) = \frac{1}{D!}\sum_{\pi=1}^{D!}\log p_{\theta}\left(x^{(\pi)}_{\Mcal(t)}|\xc^{(\pi)}_{\mathcal{R}(1:D-t)}\right).$$

Here, we can easily see that there are $\binom{D}{t-1}$ choices for the \emph{unmasked} tokens in the r.h.s.\@ of the conditional distribution, where we have previously fixed the index $t$. If we set the \emph{binomial} coefficient $\mathcal{C}_t \equiv \binom{D}{t-1}$ as the maximum number of choices, we can obtain the following equality
\begin{equation}
    \sum_{\pi=1}^{D!}\log p_{\theta}\left(x^{(\pi)}_{\Mcal(t)}|\xc^{(\pi)}_{\mathcal{R}(1:D-t)}\right) = \sum_{\pi=1}^{\mathcal{C}_t}\sum_{j=1}^{D-t+1}\log p_{\theta}\left(x^{(\pi)}_{\Mcal(j)}|\xc^{(\pi)}_{\mathcal{R}(1:D-t)}\right),
    \label{eq:perm_equal}
\end{equation}
since $D!>\mathcal{C}_t~~\forall t\in \{1,2,\dots, D\}$. Notice that once we have chosen a specific order $(\pi)$ in the masking pattern of $\Mcal$ and $\Rcal$ in Eq.~\ref{eq:perm_equal}, there are still $(D-t+1)$ choices for the \emph{masked} tokens under evaluation in the probability distribution. Alternatively, we can think of this method as taking advantage of the properties of probability to split the $D!$ choices in the order of log-conditionals into the two sums in Eq.~\ref{eq:perm_equal}. The driving idea is then that the two sums in the previous expression still remain \emph{invariant} given any $t \in \{1,2,\dots, D\}$.

Using the previous notion in Eq.~\ref{eq:masked_tokens}, we obtained our main result, which holds under the assumption of i.i.d.\@ observations with correlated tokens and the previous definition of the \textsc{lml} as the integral over the stochastic latent variables in the model.

\begin{framed}
\vspace*{-0.95\baselineskip}
\begin{proposition}
  --- The cumulative expected loss of masked pre-training along the sizes of the mask of tokens $M \in \{1,2,\dots, D\}$ is equivalent to the log-marginal likelihood of the model when using self-predictive conditionals probabilities, such that
    \begin{equation}
        \log p_{\theta}(\xc) = \sum^{D}_{m=1} \Scal_{\theta}(\xc; m),
        \label{eq:lml_main_equation}
    \end{equation}
where the score function $\Scal_{\theta}(\cdot; M)$ corresponds to
\begin{equation*}
\Scal_{\theta}(\xc; M) = \frac{1}{\Ccal_M}\sum^{\Ccal_M}_{\pi=1}\frac{1}{M} \sum^{M}_{j=1} \log p_\theta(x^{(\pi)}_{\Mcal(j)}|\xc^{(\pi)}_{\mathcal{R}(1:D-j)}) = \frac{1}{M} \mathbb{E}_{\Mcal}\left[\sum^{M}_{j=1} \log p_{\theta}(x_{\Mcal(j)}|\xc_{\Rcal})\right].
 \label{eq:s_definition}
 \end{equation*}
Proof: In the supplementary material.
 \end{proposition}
 \vspace*{-0.5\baselineskip}
\end{framed}
It is remarkably important to link the sum of log-conditionals $\log p_{\theta}(x_{\Mcal(j)}|\xc_{\Rcal})$ in our proposition with the main objective used in \textsc{mpt} in Eq.~\ref{eq:mpt_loss}. The main message of our result is that the \emph{score function} $\Scal_{\theta}(\cdot; t)$  acts as an average over the different random masks. These shape the structure of conditioning in probabilities. The cumulative sum of the score function $\Scal_{\theta}(\cdot; t)$ over different sizes of the \textsc{mpt}'s mask formally leads to the true value of the model's \textsc{lml}. This result is exact whenever we consider the closed-form self-predictive probabilities of the model and \emph{all} the possible choices for the masking pattern $\Mcal$. Since this is usually not affordable, due to the combinatorial cost and the lack of tractability, we usually have a \emph{biased} estimator. However, it is still sufficient to prove that \textsc{mpt} maximizes \textsc{lml} during training as we will show later. This point will be discussed in the following empirical studies. Further details on the derivations are included in the supplementary material.

\subsection{Formal results in tractable models}
To verify that masked pre-training effectively maximizes \textsc{lml}, we need a tractable probabilistic model based on latent variables as the \emph{proof-of-concept}. Probabilistic \textsc{pca} (\textsc{ppca}) \citep{tipping1999probabilistic} is perhaps the best option here, as it has been previously used to understand other empirical observations in generative methods, e.g.\@ posterior \emph{collapse} in \textsc{vae}s \citep{lucas2019don}, or even considered as the starting point of \textsc{gplvm}s \citep{lawrence2005probabilistic}. In particular, the \textsc{ppca} model assumes that Gaussian observations map linearly to sets of real-valued latent variables $\zc_{1:n}$, such that $\xc = \Wb \zc + \bm{\mu} + \epsilon$, where $\epsilon \sim \Ncal(0, \sigma^2_0 \mathbb{I})$. Importantly, the prior is conventionally defined as isotropic, where $p(\zc) = \Ncal(0,\bm{1})$. We are therefore interested in the closed form expression of the \textsc{ppca}'s $\textsc{lml}$, which also factorizes across samples as follows
\begin{equation}
    \log p_\theta(\xc_{1:n}) = \sum_{i=1}^{n}\log p_\theta(\xc_{i}),~~~~~~~~\text{where}~~~ p_\theta(\xc_{i}) = \Ncal(\xc_i|\bm{\mu}, \Sb),
\end{equation}
and we obtain the covariance matrix using $\Sb = \Wb\Wb^{\top} + \sigma^2_0 \mathbb{I}$. For our analysis, the Gaussian nature of $p_\theta(\xc_{i})$ is of fundamental importance. Given the random mask $\Mcal$, the self-predictive conditionals used in \textsc{mpt} naturally emerge from the formulation using properties of Gaussian marginals, such that $p_{\theta}(\xc_{\Mcal}|\xc_{\Rcal}) = \Ncal(\bm{m}_{\Mcal|\Rcal}, \bm{v}_{\Mcal|\Rcal})$ is parameterized according to
\begin{equation}
\bm{m}_{\Mcal|\Rcal} = \Sb^{\top}_{\Mcal\Rcal}\Sb^{-1}_{\Rcal\Rcal} \xc_{\Rcal}, ~~~~\bm{v}_{\Mcal|\Rcal} = \Sb_{\Mcal\Mcal} + \Sb^{\top}_{\Mcal\Rcal}\Sb^{-1}_{\Rcal\Rcal} \Sb_{\Mcal\Rcal},
\end{equation}
where we split the \textsc{lml} covariance matrix $\Sb$ into the blocks corresponding to the indices included in $\Mcal$ and $\Rcal$. We use these \emph{mean} and \emph{variance} parameters of the self-predictive density to recursively evaluate the log-probabilities in Prop.~1. In practice, two elements become critical for the computation, one is the size $M$ of the \emph{mask} and another is the number of random masks $P<\Ccal_{M}$ considered. These induce a \emph{trade-off} between accuracy and computational cost. Moreover, their role in approximating \textsc{lml} using a biased estimate is carefully analyzed in the following empirical studies. We also included additional details on the previous derivations in the supplementary material.

\paragraph{Fast asymptotic convergence.}
Our theory indicates that we should evaluate \emph{all} $\Ccal_{M}$ random masks of the tokens to achieve the exact value of the \textsc{lml}. However, even if the combinatorial nature of the sum in the r.h.s.\@ of the last equation in Prop. 1 becomes very large when the dimensionality of data augments, we suspect that it might converge relatively fast to the true value of the \textsc{lml}. This hypothesis would explain why large models that are fitted with standard \textsc{mpt} generalize well using just one random mask per training epoch.

\begin{figure}[ht!]
	\centering
	\includegraphics[width=1.0\textwidth]{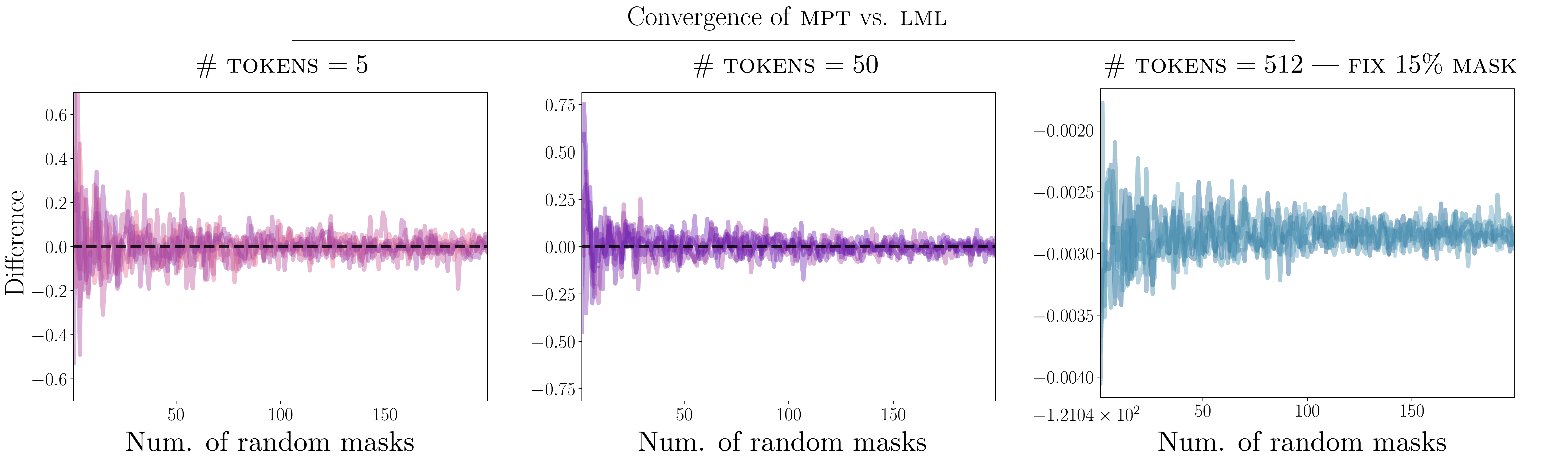} 
	\caption{Asymptotic convergence of the cumulative \textsc{mpt} loss to \textsc{lml} as the number of random masks $P$ augments. Curves indicate the relative difference, where $0.0$ means that $\textsc{mpt}$ equals $\textsc{lml}$. \textbf{(Left).}  Each observation consists of 5 tokens. \textbf{(Center)} Each observation consists of 50 tokens. \textbf{(Right).} Observations have $512$ tokens and the rate of masking is fixed to $15\%$ (76 tokens). As the theory indicates, when the size of $\Mcal$ is fixed, the cumulative \textsc{mpt} loss becomes a \emph{biased} estimator of the \textsc{lml}. The curves converge asymptotically to the bias.}
    \label{fig:convergence}
\end{figure}

Here, we empirically study if the cumulative \textsc{mpt} loss converges to the \emph{true} value of the \textsc{lml} under the definition of the \textsc{ppca} model. In particular, to the \textsc{lml} obtained with the original parameters that generated the data. The results in Fig.~\ref{fig:convergence} and Tab.~\ref{tab:metrics} indicate that as long as we average over more random masking patterns, the cumulative \textsc{mpt} loss approximates the \textsc{lml} of the model very well. Thus, having defined a \textsc{ppca} model with a latent space of $K=2$ dimensions, we observe in the \emph{left} and \emph{middle} plots that the asymptotic convergence happens for both small ($D=5$) and large ($D=50$) number of tokens per observation. Additionally, we observe that the estimation of \textsc{lml} is clearly \emph{unbiased} if we use the cumulative \textsc{mpt} loss according to Eq.~\ref{eq:s_definition}, which is an important insight. Notice that $P=1$ is usually set up in \textsc{mpt} in practice.
\begin{table}[h!]
        \centering
        \caption{Evolution of negative \textsc{mpt} loss w.r.t.\@ max.\@ number of random masks $P$.}
	\resizebox{0.65\textwidth}{!}{
		\color{black}\scriptsize
		\begin{tabular}{cccc}
			\toprule
			\textsc{True \textsc{lml} ($\uparrow$)}  & \textcolor{black}{$P=1$} & \textcolor{black}{$P=10$} & \textcolor{black}{$P=100$} \\
			\midrule
			$(-60.34)$ & $-60.44 \pm 0.47$ & $-60.22 \pm 0.12$ & $\maxf{-60.34 \pm 0.03}$  \\
			\bottomrule
		\end{tabular}
		\label{tab:metrics}
	}
\end{table}
Additionally, we tested the tractable model using a dimensionality similar to the input data used in \textsc{bert} \citep{devlin2018bert}, where the number of tokens is typically $D=512$ per observation and the mask rate is fixed to $15\%$. The fact of fixing the rate of masking in \textsc{mpt} produces that the sum in Eq.~\ref{eq:lml_main_equation} is incomplete. Thus, we have a \emph{biased} estimation of the \textsc{lml}. However, this bias is \emph{known} and constant during the training of parameters $\theta$, which does not prevent the general maximization of \textsc{lml}. This point is carefully analyzed in the next empirical study with learning curves. One additional finding here is that as $P\rightarrow \Ccal_{M}$, the cumulative \textsc{mpt} loss also converges asymptotically to the \emph{biased} estimator of the \textsc{lml} as shown in the right plot in Fig.~\ref{fig:convergence}.

\begin{figure}[ht!]
	\centering
	\includegraphics[width=1.0\textwidth]{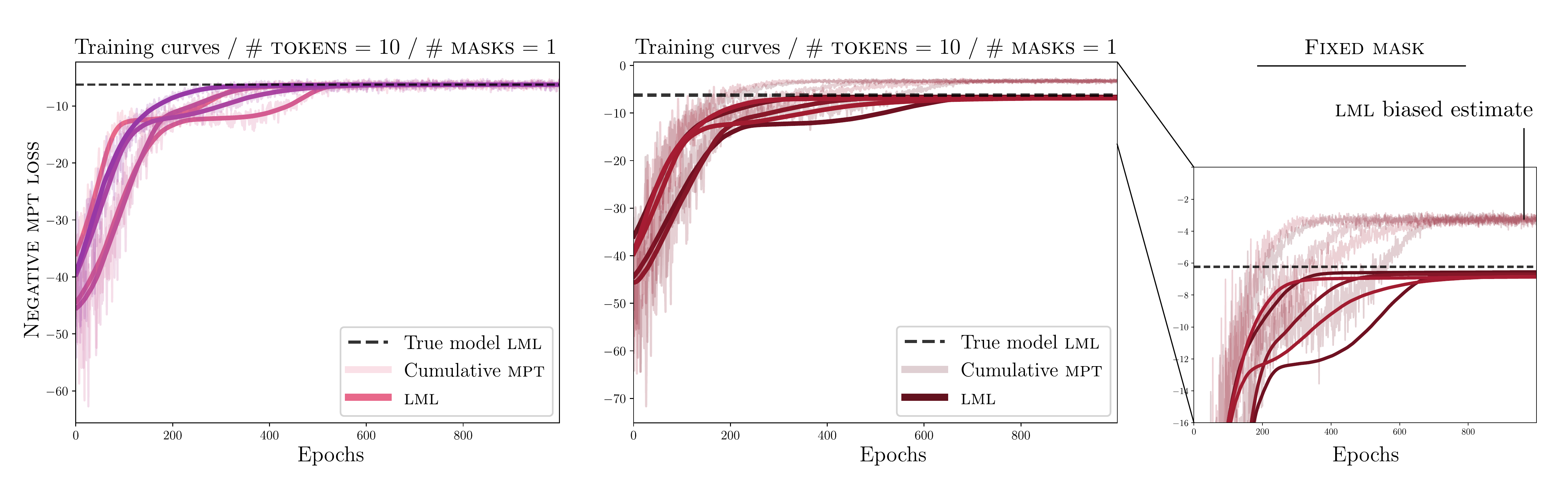} 
	\caption{Training curves of the negative cumulative \textsc{mpt} loss in \textsc{ppca} vs.\@ the ground truth (\textsc{gt}) \textsc{lml}. The number of samples is $N=2000$ and the number of tokens is $D=10$. All plots used $P=1$ random masks per epoch and five different initializations. \textbf{(Left).} The rate of masking is \emph{unfixed} and it varies from $1\%$ until $100\%$. The negative \textsc{mpt} loss converges to the \textsc{gt}-\textsc{lml} (dashed line). Darker curves are the exact \textsc{lml} per epoch. \textbf{(Center).} Convergence with \emph{fixed} mask to $20\%$ of tokens. The negative \textsc{mpt} loss is no longer centered around the \textsc{lml} and it converges to a \emph{biased} estimate. \textbf{(Right).} \emph{Zoomed} curves of convergence. The \emph{bias} is constant and all \textsc{mpt} losses converge to the same point. The \textsc{lml} per epoch is also maximized and converges to \textsc{gt}-\textsc{lml}.}
    \label{fig:training_curves}
\end{figure}

\paragraph{LML maximization and biased estimation.} We next seek to extend the previous study to understand the behavior of the cumulative \textsc{mpt} loss in training curves. So far, we have observed how the number of random mask patterns affects the precision around the \emph{unbiased} estimation of the \textsc{lml}. Theory and previous empirical results indicate that we are targeting \textsc{lml} or at least a decent \emph{biased} estimate of \textsc{lml} when averaging self-predictive conditionals as in \textsc{mpt}. However, we still want to examine if this maximizes \textsc{lml} in \emph{all cases} and under stochastic gradient optimization. This principal hypothesis is confirmed in Fig.~\ref{fig:training_curves}, where different training curves are shown for different initializations and setups of the same \textsc{ppca} model. The key insight showed by this experiment is that the exact \textsc{lml} is iteratively maximized at each epoch, in parallel with the maximization of the negative \textsc{mpt} loss. On the other side, we also have that \textsc{mpt} is an unbiased stochastic approximation of \textsc{lml}, as in Fig.~\ref{fig:convergence}, whenever we consider different rates of random masking $\Mcal$. We can also observe that as soon as we fix the size of the mask to index the $20\%$ of tokens, the \textsc{mpt} loss becomes a \emph{biased} estimate. Intuitively, this is equivalent to fixing $M$ in the sum in Eq.~\ref{eq:lml_main_equation}. Again, it converges to the same value from different initializations of parameters $\theta$. Additionally, we highlight that the \textsc{lml} is still maximized in this case, which is of high similarity to practical uses in larger models. Overall, this result first confirms the main insight of the work on the link between generalization when using \textsc{mpt} and the maximization of the model's \textsc{lml}.

\begin{figure}[ht!]
	\centering
	\includegraphics[width=0.85\textwidth]{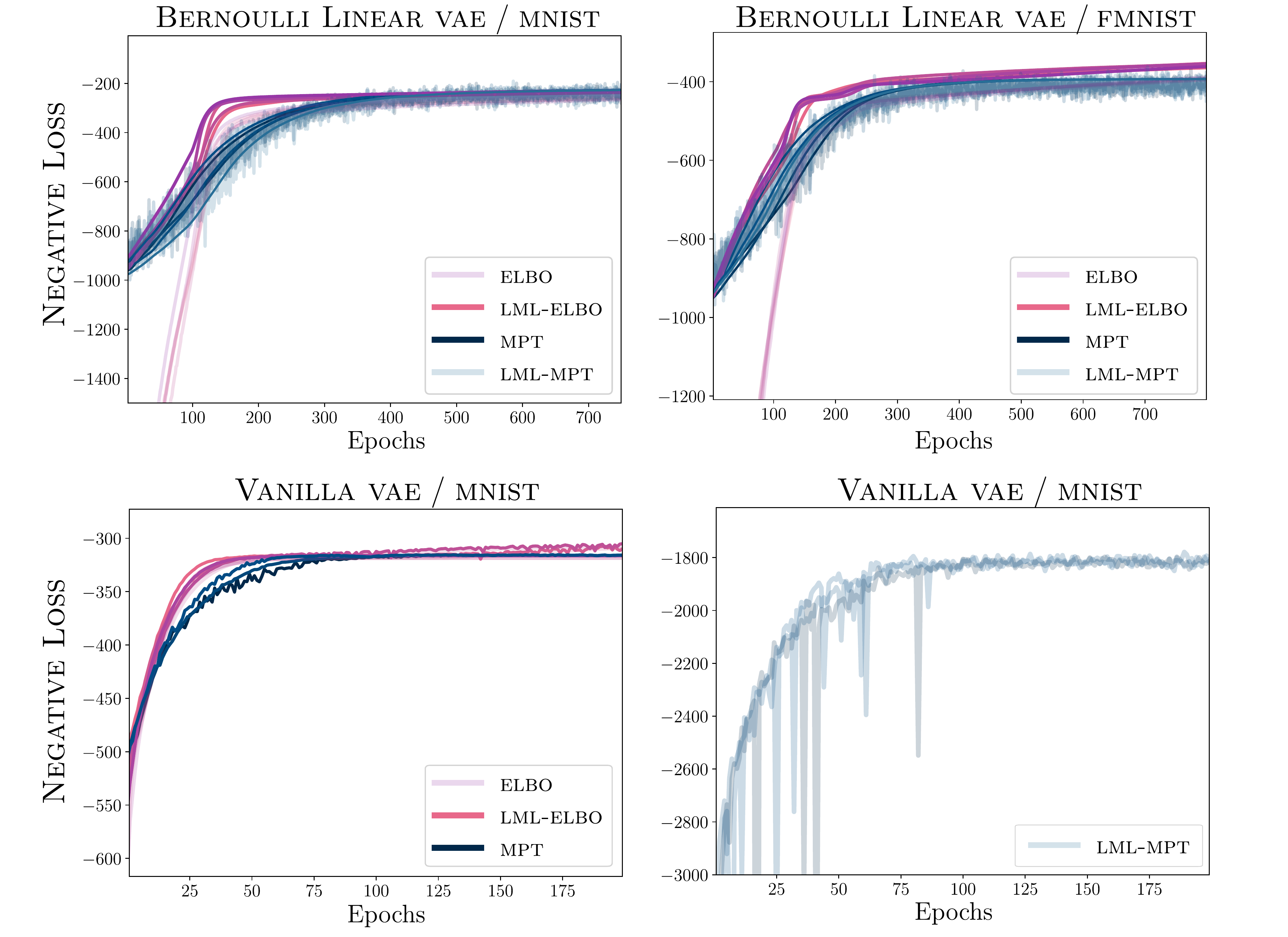} 
	\caption{Training curves for linear \textsc{vae} and deep \textsc{vae} models with variational inference (\textsc{vi}) and \textsc{mpt}. Data consist of subsets of \textsc{mnist} and \textsc{fmnist}. \textbf{(Upper Row).} A linear \textsc{vae} model with Bernoulli likelihood function in $N=2000$ samples of \textsc{mnist} and \textsc{fmnist}. Shaded curves correspond to the target losses used in the optimizer (\textsc{elbo} and \textsc{mpt}). Darker lines indicate the evolution of the \textsc{lml}, which are approximated via numerical integration in a latent space $\Zcal$ of dimensionality $K=2$. \textbf{(Lower Row).} Vanilla \textsc{vae} with Gaussian likelihood for \textsc{mnist}. The \textsc{lml} curves are approximated via Monte Carlo (\textsc{mc}) samples. Self-predictive conditional probabilities are obtained via \emph{recursive} encoding-decoding. The size of the random masking is fixed and set to $33\%$.}
        \label{fig:elbo_curves}
\end{figure}
\begin{figure}[ht!]
	\centering
	\includegraphics[width=1.0\textwidth]{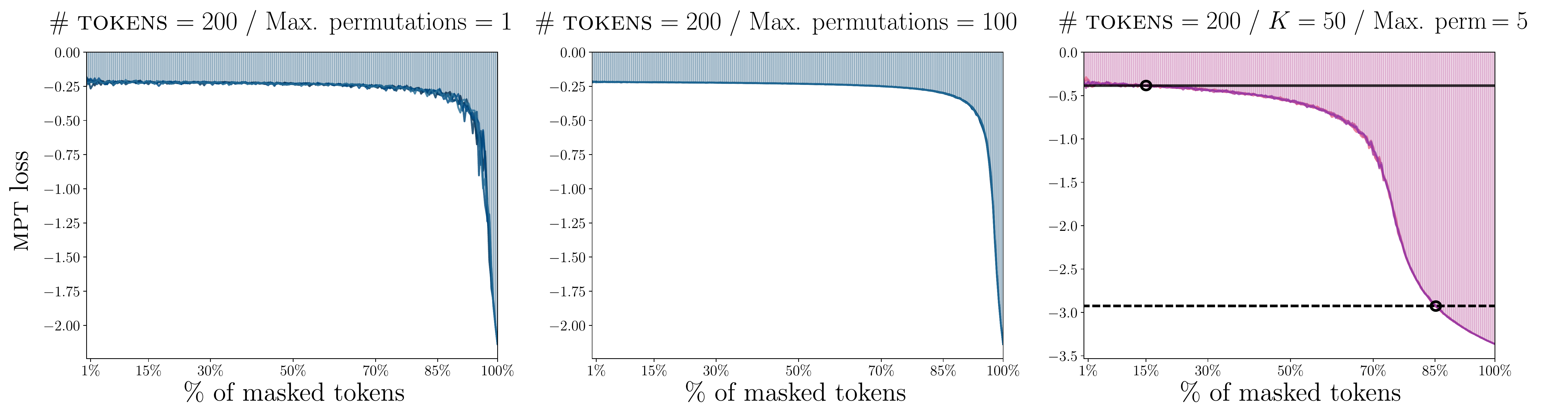} 
	\caption{Area under the curve described by $\Scal_\theta(\cdot; M)$. The area is approximately equal to the model's \textsc{lml} according to the theory. Larger probability values are obtained with smaller rates of masking. \textbf{(Left).} Area described with $P=1$ random masking per epoch. The curve is more noisy and the area slightly looses precision w.r.t.\ \textsc{lml}. \textbf{(Center).} Area under the \textsc{mpt} curve for $P=100$. \textbf{(Right).} Latent space is augmented to be of $K=50$. Decay of predictive probabilities begins at around $50\%$ masking rate.}
    \label{fig:areas}
\end{figure}
\paragraph{Beyond tractable models and implicit integration.} One remaining question in our analysis is how the probabilistic theory around \textsc{mpt} adapts to intractable or non-linear models. In practice, self-predictive probabilities imply integrating out the latent variables, often given the posterior distribution. In most cases, performing this integration is extremely difficult or not possible in training time. Therefore, we are interested in finding if alternative approximations $q_{\theta}$ to the \emph{true} self-conditional probabilities still produce accurate estimation and maximization of the \textsc{lml}. This point is confirmed in Fig.~\ref{fig:elbo_curves}. Inspired by the experiments of \citet{lucas2019don} with \emph{linear} \textsc{vae}s, we set up a Bernoulli likelihood on top of the latent variable model. The tractable formulation in the Gaussian example coincides with \textsc{ppca}. Since predictive conditionals are no longer tractable for us, we use numerical integration to obtain the probabilities of masked tokens. In Fig.~\ref{fig:elbo_curves}, we test the training with the cumulative \textsc{mpt} loss as well as we compare with standard variational inference using the model's evidence lower bound (\textsc{elbo}). For the \emph{mini}-dataset with \textsc{mnist} samples, we observe that both models converge to a similar value of the \textsc{lml}. Thus, the fundamental insight here is that \textsc{mpt} maximizes \textsc{lml} even under training with approximate self-predictive conditional probabilities. For the \textsc{lml} curves, we also used numerical integration.

Beyond linear models, our theory is useful when applied to non-linear models. Moreover, in Fig.~\ref{fig:elbo_curves} we also include the results for \emph{deep} \textsc{vae}s based on \textsc{nn}s. While the estimation of \textsc{lml} was obtained via Monte Carlo (\textsc{mc}) samples, we used iterative \emph{encoding}-\emph{decoding} to produce the self-conditional probabilities for masked tokens --- see Sec.~F in \citet{rezende2014stochastic}. In this scenario, we also observe the maximization of the \textsc{lml} according to the evolution of the \textsc{mpt} loss.

Another key insight showed by this study is the ability of \textsc{mpt} to perform \emph{implicit} integration. The cumulative sum over the different rates of random masking is another way to see a discrete integral under the curve described by the score function $\Scal_\theta(\cdot; M)$ in Eq.~\ref{eq:lml_main_equation}. In Fig.~\ref{fig:areas}, we show the areas under the curve and the effect of reducing the number of random masks $P$. The blue plots correspond to a trained \textsc{ppca} model and the area corresponds to the \textsc{lml} estimate. The long tail in the right part of the curves, when the rate of masking is larger than $90\%$, indicates that the model is no longer able to produce good estimates of the tokens with only $10\%$ of the input dimensions observed. This explains, why the probabilities have an approximately exponential decay. However, this effect is not constant, and it might depend on the latent structure of the model. In the r.h.s.\ plot we observe that the decay of conditional probabilities happens earlier at approximate $50\%$ random masking or larger. The role of the masking rate is perhaps the missing part in the picture \citep{wettig2022should}, as it is the one that determines the approximation to the \textsc{lml}. With the purpose of providing an intuition on how rates of $15\%$ or $85\%$ effect to the area under the curve, we indicate with two black lines the approximate area that approximates the \textsc{lml}. A longer discussion is provided in the supplementary material. 

\subsection{Applied theory on large language models}

\begin{figure}[ht!]\label{fig:4}
	\centering
	\includegraphics[width=1.0\textwidth]{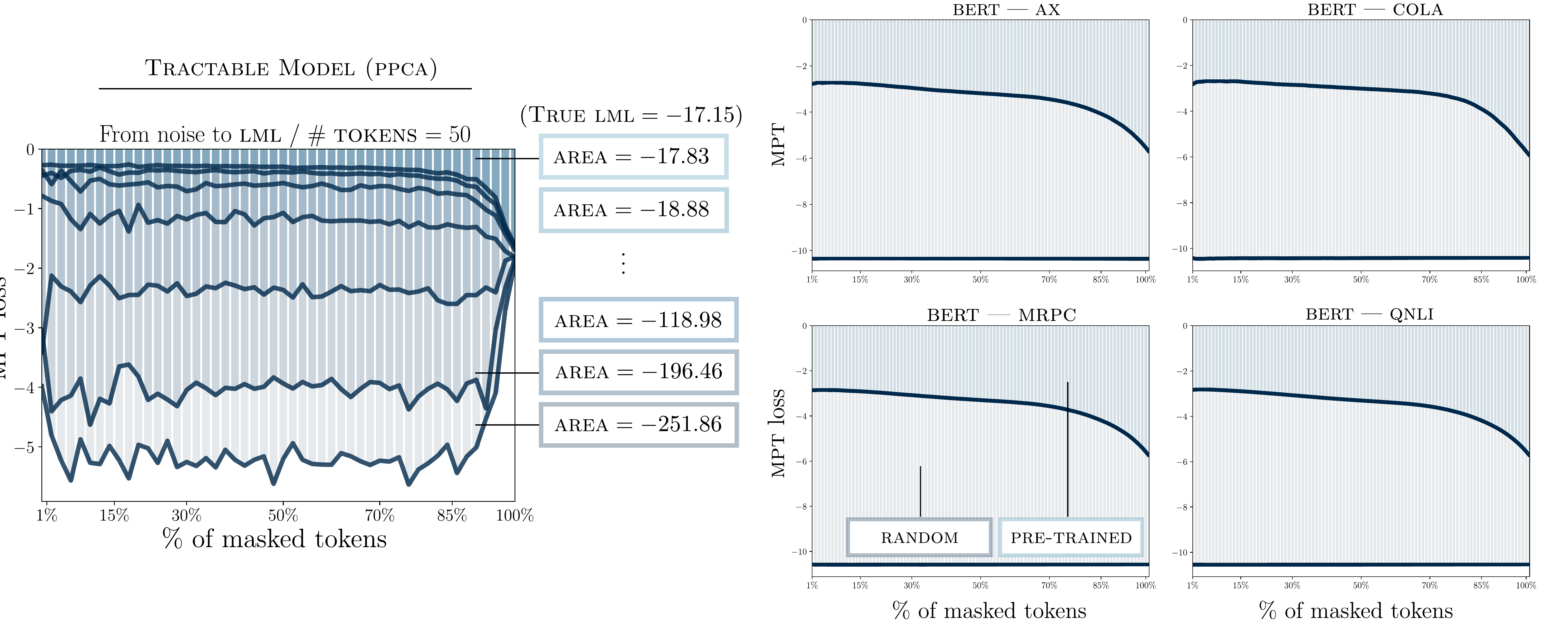} 
        \vspace*{-0.5\baselineskip}
	\caption{Evolution of  the area under the \textsc{mpt} curve. Comparison between one tractable model (\textsc{ppca}) and \textsc{bert}. The area under the curves is \emph{approximately} the \textsc{lml}. Random initialization of the parameters produces \textsc{mpt} curves with similar low probabilities for all $\%$ of masking. As the number of epochs increases, the curve brings higher values of log-probability for lower ratios of masking. The area also converges to the true value of \textsc{lml}. \textbf{(Left).} \textsc{ppca} model trained for $600$ epochs. Each curve represents $\{0, 100, 200, 300, 400, 500, 600\}$ epochs of training with \textsc{mpt}. \textbf{(Right).} Random initialization and end-of-pretraining curves for the \textsc{mpt} loss w.r.t.\@ the $\%$ of masked tokens. Curves are similar but not identical for the $4$ different datasets given the pre-trained \textsc{bert} model.}
    \label{fig:llms}
 \vspace*{-0.5\baselineskip}
\end{figure}

In this section, we aim to understand how the area under the \textsc{mpt} curve evolves and behaves for large language models (\textsc{llm}s). While the direct computation of the \textsc{lml} is not feasible for non-linear transformer models, we are interested in checking how the rate of masking affects the curve compared with the tractable \textsc{ppca} model. The results provided in Fig.~\ref{fig:llms} and Tab.~\ref{tab:llms} give us insights into this behavior. First, we observe that the \textsc{mpt} curve is approximately \emph{flat} for every rate of masking in the \textsc{ppca} when parameters are randomly initialized. Intuitively, this indicates that the model is not able to correctly predict any token given some context. In some way, it produces noise independently of the number of conditional tokens, which explains the low log-probabilities. Second, we can also notice that the curve changes its shape as more training epochs are considered. The curve after $600$ epochs produces high probability values for different rates of masking, while the long tail of low probabilities appears when masking more than $85\%$ of tokens. Moreover, the area under these curves is the estimation of the \textsc{lml}, which accurately converges to the \emph{ground truth} value of the \textsc{lml} with the original generative parameters.

For the study of the curves in \textsc{llm}s, we used four datasets from the General Language Understanding Evaluation (\textsc{glue}) \citep{wang2019glue}. Additionally, we consider a $110$M parameters \textsc{bert} model using pre-trained checkpoints\footnote{Pre-trained parameters for the \textsc{bert} model are available in the library --- \url{https://huggingface.co/}.} and random initializations. To draw the \textsc{mpt} curves, we computed the mean cross-entropy per each rate of masking between $1\%$ and $99\%$. In Fig.~\ref{fig:llms}, we observe that random initializations of \textsc{bert} parameters lead to \emph{flat} curves of low self-predictive probabilities. On the other hand, the pre-trained curves show a similar behavior as in the tractable model, where the area is reduced and a longer tail of low probabilities happens when the rate of masking becomes larger. This result supports our hypothesis that \textsc{mpt} in \textsc{llm}s might be performing implicit integration of the latent space and maximising the marginal likelihood of the model.

\begin{table}[ht!]
        \caption{Area under the \textsc{mpt} curve for \textsc{bert} model and four \textsc{glue} datasets.}
        \centering
	\resizebox{0.75\textwidth}{!}{
		\color{black}\scriptsize
		\begin{tabular}{ccccc}
			\toprule
			  \textsc{Glue datasets} $\rightarrow$ &\textsc{ax} & \textsc{cola} & \textsc{qnli} & \textsc{mrpc}\\
			\midrule
			\textsc{Area / Random init.} ($\uparrow$) & $\maxf{-5245.31}$ & $-5283.52$ & $-5343.98$ & $-5362.21$ \\
                \textsc{Area / Pre-trained} ($\uparrow$) & $-1715.75$ & $\maxf{-1657.68}$ & $-1770.28$ & $-1773.45$\\
			\bottomrule
		\end{tabular}
		\label{tab:llms}
	}
\end{table}

\textbf{Reproducibility.} All the empirical studies and results are \emph{reproducible}. We provide the code and details for every figure in the public repository at \texttt{https://github.com/pmorenoz/MPT-LML/}.

\vspace*{-0.25\baselineskip}
\section{Related work}
\vspace*{-0.25\baselineskip}
Masked pre-training and large scalability are key elements of the current success of transformer models \citep{vaswani2017attention} on natural language processing (\textsc{nlp}) tasks. Vision transformers (\textsc{v}i\textsc{t}) \citep{dosovitskiy2020image} bridged the architectural gap between  \textsc{nlp} and computer vision, making masked language modeling (\textsc{mlm}) suitable for images. In this regard,  \textsc{b}ei\textsc{t} \citep{bao2022beit} adopted  \textsc{v}i\textsc{t} and proposed to mask and predict discrete visual tokens. Most recently, masked autoencoders \citep{he2022masked} also adopted masked pre-training by predicting pixel values for each masked patch, and \textsc{be}i\textsc{t3} \citep{wang2022image} performs \textsc{mlm} on texts, images, and image-text pairs, obtaining state-of-the-art performance on all-vision and vision-language tasks. Additionally, masked pre-training has been successfully adapted to video \citep{tong2022videomae}, where random temporal cubes are iteratively masked and reconstructed.

The susprising ability of recent generative models to generalize and do impressive in-context learning has inspired earlier works to study this phenomena from the Bayesian lens. The notion that \textsc{llm}s might be performing \emph{implicit} Bayesian inference was first described in \citet{xie2021explanation} where in-context learning is described as a mixture of \textsc{hmm}s. However, the equivalence between the log-marginal likelihood and \emph{exhaustive} cross-validation was first provided in \citet{fong2020marginal}. Earlier works \citep{vehtari2002bayesian,gelman2014understanding} also provided a Bayesian perspective of \textsc{cv}. Additionally, \citet{pablo:neurips:2022} leveraged this link for training Gaussian process models according to a stochastic approximation to the marginal likelihood. Similarly to current masked pre-training, the size of the conditioning variable (masking rate) was held constant. This was reported to improve notably upon traditional variational lower bounds.

\vspace*{-0.25\baselineskip}
\section{Discussion and outlook} 
\vspace*{-0.25\baselineskip}
In this paper, we have shown that masked pre-training implicitly performs stochastic maximization of the model's marginal likelihood. The latter is generally acknowledged as being an excellent measure of a model's ability to generalize \citep{fong2020marginal}, and our results help to explain the strong empirical performance associated with masked pre-training. We have further seen that the developed theory matches the empirical training behavior very well. Moreover, we illustrated the role that the rates and the number of random samples of masking play in the estimation of the \textsc{lml}. We have also provided insights and a new perspective to study masked pre-training in tractable models while also finding strong similarities with \textsc{llm}s.

\paragraph{Limitations.} We have developed a formal probabilistic theory which links masked pre-training with the Bayesian principles. While we provide evidence that generalization in recent large models is related to the maximisation of the marginal likelihood, these methods usually introduce new elements that improve performance but may not entirely fit to the propositions provided in this work. In practice, this is not a limitation but a remark that there is still room for understanding the abilities of recent generative modelling. On this regard, one example might be autoregressive modeling between the masked tokens. While these are not currently analyzed in our work, we hypothesize that they could also be linked in further development to our formal propositions.

\paragraph{Relevance for large models using masked pre-training.}
We have shown empirical results of the connection between \textsc{mpt} and \textsc{lml}. This link sheds light on the understanding of generalization, particularly in recent pre-trained models. One positive outcome of our studies is the notion of having \emph{biased} Bayesian estimators whenever a practitioner fixes the masking rate, e.g.\ to $15\%$. Currently, there is a significant interest around the role of masking rates in \textsc{llm}s \citep{wettig2022should}. These studies could benefit from the insights provided in this paper. We also argue that the theory provides \emph{hints} that may be beneficial, for instance, for uniformly sampling the mask size, instead of the current fixed-rate practice.

\paragraph{Relevance for Bayesian models.}
Current Bayesian modeling is dominated by approximate methods. Variational inference foregoes the ambition of training according to the marginal likelihood and instead resorts to bounds thereof. This inherently yields suboptimal models. Our theory suggests that if we can design Bayesian models in which conditioning is \emph{cheap}, then we can stochastically optimize w.r.t.\ the true marginal likelihood easily. Beyond shedding light on the success of masked pre-training, theory also suggests that large-scale Bayesian models could be successfully trained in the future with appropriately designed self-supervision.

\small
\bibliography{biblio}

\newpage

\begin{center}
    \begin{Large}
        \texttt{Appendix}
    \end{Large}
\end{center}
\noindent\makebox[\linewidth]{\rule{\textwidth}{0.4pt}}

\vspace{0.5cm}

In this appendix, we provide additional details about the theoretical and empirical results included in the manuscript. These indicate that masked pre-training optimizes according to a stochastic gradient of the model's log-marginal likelihood. We remark that our main proof relies on a previous observation from \citet{fong2020marginal}, who shows that log-marginal likelihood (\textsc{lml}) is equivalent to an \emph{exhaustive} cross-validation score over \emph{all} training-test data partitions in the dataset. While \citeauthor{fong2020marginal}' formal proof uses properties of probability to link cross-validation on observations with marginal likelihood, we use them to prove that self-conditional probabilities across masked features also lead to the model's \textsc{lml}. Additionally, we include the code for experiments and extra details on the tractable linear model as well as the initial setup of hyperparameters for the reproducibility of our results.

\section*{A~~~Proof of Proposition}

\emph{Proof.} Consider an i.i.d.\ dataset $\xc_{1:n}$, where each $i^{\text{th}}$ object $\xc_{i}\in \mathcal{X}^{D}$ for some continuous or discrete domain $\mathcal{X}$. We define a \emph{latent variable model} where the likelihood function is defined as $p_{\theta}(\xc|\zc)$, where $\zc \in \mathcal{Z}^{K}$ are the latent objects for some domain $\mathcal{Z}$ and the prior distribution is $p(\zc)$. These assumptions lead us to a log-marginal likelihood (\textsc{lml}) of the model that factorises across observations, such that $\log p_\theta(\xc_{1:n}) = \sum^{n}_{i=1}\log p_\theta(\xc_{i})$, where  $p_\theta(\xc_{i}) = \int p_{\theta}(\xc_{i}|\zc_{i})p(\zc_{i})\mathrm{d}\zc_{i}$.

Using the properties of probability, we can rewrite the \textsc{lml} of each $i^{\text{th}}$ object as a sum of conditional distributions between dimensions or \emph{features}. This sum is of the form
\begin{equation}
    \log p_{\theta}(\xc) = \sum^{D}_{t=1}\log p_{\theta}\left(x_t|\xc_{t+1:D}\right), 
    \tag{A.1}
    \label{eq:lml_sum}
\end{equation}
where we omitted the $i^{\text{th}}$ subscript to keep the notation uncluttered. Here, we see that the value of $\log p_{\theta}(\xc)$ is \emph{invariant} to the choice of the conditional probabilities in \eqref{eq:lml_sum} if these ones follow the \emph{chain-rule} of probability according to the $D$ dimensions of $\xc$. Additionally, this indicates that we have $D!$ different choices for the sum of conditional probabilities in \eqref{eq:lml_sum}, which allows us to write
\begin{equation}
    \log p_{\theta}(\xc) = \frac{1}{D!}\sum^{D!}_{\pi=1}\sum^{D}_{t=1}\log p_{\theta}\left(x^{(\pi)}_{\Mcal(t)}|\xc^{(\pi)}_{\Mcal(t+1:D)}\right).
    \tag{A.2}
    \label{eq:lml_ave}
\end{equation}
Here, we defined $\mathcal{M}$ as the \emph{indexing mask}, which consists of indices drawn from $\{1,2,\dots, D\}$, and we initially assume in \eqref{eq:lml_ave} that $|\Mcal| = D$. The $\pi^{\text{th}}$ superscript indicates the \emph{order} of indices used to produce the conditional chain-rule.

If we then swap the order of sums in \eqref{eq:lml_ave} and we fix the index $(t)$ in \eqref{eq:lml_ave}, we can see that there are $(D-t+1)$ choices for the \emph{tokens} under evaluation by the probability distribution and $\binom{D}{t-1}$ choices for the rest of conditional factors. We can then write 
\begin{equation*}
    \sum^{D!}_{\pi=1}\log p_{\theta}\left(x^{(\pi)}_{\Mcal(t)}|\xc^{(\pi)}_{\Mcal(t+1:D)}\right) = \sum^{\mathcal{C}_t}_{\pi=1}\sum^{D-t+1}_{j=1}\log p_{\theta}\left(x^{(\pi)}_{\Mcal(j)}|\xc^{(\pi)}_{\Mcal(t+1:D)}\right).
\end{equation*}
To match notation with masked pre-training (\textsc{mpt}), we set $\Mcal$ to be the \emph{masked} subset of indices sampled from $\{1,2,\dots, D\}$, such that $M<D$ and the rest of \emph{unmasked} indices shape the complementary subset $\Rcal = \{1,2,\dots, D\} \setminus \Mcal$. Using the previous sum and notation in \eqref{eq:lml_ave}, we can finally state that the $\textsc{lml}$ is a \emph{cumulative} sum of averages, such that
\begin{equation}
    \log p_{\theta}(\xc) = \sum^{D}_{t=1}\frac{1}{\mathcal{C}_t}\sum^{\mathcal{C}_t}_{\pi=1}\frac{1}{D-t+1}\sum^{D-t+1}_{j=1}\log p_{\theta}\left(x^{(\pi)}_{\Mcal(j)}|\xc^{(\pi)}_{\Rcal(1:D-t)}\right).
    \tag{A.3}
\end{equation}
Setting $M=D-t+1$ and rearranging $\mathcal{C}_{t}$ as $\mathcal{C}_{M} = \binom{D}{D-M}$ gives us the formal result included in Proposition 1.

\section*{B~~~Full view of Probabilistic PCA}

Probabilistic \textsc{pca} (\textsc{ppca}) \citep{tipping1999probabilistic} is a \emph{latent variable model} in which the marginal likelihood distribution is tractable and the maximum likelihood solution for the parameters can be analytically found. The model also assumes that the data are $D$-dimensional observations $\xc$. Additionally, we assume that there exists a low-dimensional, where each sample has a \emph{latent} representation $\zc$$\in$$\Zcal$ for each datapoint, where $\Zcal=\mathbb{R}^{K}$. The relationship between the latent variables and the observed data is \underline{linear} and can be expressed as
$$\xc = \Wb \zc + \bm{\mu} + \epsilon,$$
where $\epsilon \sim \Ncal(0,\sigma^{2}_0\mathbb{I})$, $\bm{\mu} \in \mathbb{R}^{D}$ and $\Wb\in\mathbb{R}^{D\times K}$. The likelihood model for observations $\xc$ can be then written as 
$$ p(\xc | \zc, \Wb, \bm{\mu}, \sigma^2_0) = \Ncal(\Wb \zc + \bm{\mu}, \sigma^{2}_0\mathbb{I}),$$
and more importantly, it allows the integration of the latent variables in closed-form. Thus, we can obtain the following marginal likelihood per datapoint in an easy manner
$$ p_{\theta}(\xc) = p(\xc | \Wb, \bm{\mu}, \sigma^2_0) = \Ncal(\xc| \bm{\mu}, \Wb\Wb^{\top} + \sigma^2_0\mathbb{I}),$$
where we used $\theta = \{\Wb,\bm{\mu},\sigma^2_0\}$. Moreover, under the independence assumption taken in \textsc{ppca} across $n$ observations $\xc_{1:n}$, the \emph{global} log-marginal likelihood of the model can be expressed using the following sum
$$ p_{\theta}(\xc_{1:n}) = \prod_{i=1}^{n} p_{\theta}(\xc_i).$$
\paragraph{Posterior predictive probabilities.} The predictive distribution between the dimensions of $\xc_i$ can be obtained from both latent variable integration or by properties of Gaussian conditionals. In our case, we use the latter example. Thus, having both \emph{mask} $\Mcal$ and \emph{rest} $\Rcal$ indices according to our previous notation, we can look to the multivariate normal distribution $p_{\theta}(\xc)$ using \emph{block} submatrices, such that
$$ p_{\theta}(\xc) = \Ncal\left( \begin{bmatrix}\xc_{\Mcal}\\ \xc_{\Rcal}\end{bmatrix} \Big| \begin{bmatrix}\bm{\mu}_\Mcal \\ \bm{\mu}_\Rcal \end{bmatrix}, \begin{bmatrix}\Sb_{\Mcal\Mcal} & \Sb_{\Mcal\Rcal}  \\ \Sb^{\top}_{\Mcal\Rcal} & \Sb_{\Rcal\Rcal} \end{bmatrix}\right),$$

where we also defined $\Sb = \Wb\Wb^{\top} + \sigma^2_0\mathbb{I}$. Using the properties of conditional probabilities on normal distributions, we can write the posterior predictive densities in closed-form, such that $p_{\theta}(\xc_{\Mcal}|\xc_{\Rcal}) = \Ncal(\bm{m}_{\Mcal|\Rcal}, \bm{v}_{\Mcal|\Rcal})$, where parameters are obtained from
\begin{equation*}
\bm{m}_{\Mcal|\Rcal} = \bm{\mu}_\Mcal + \Sb^{\top}_{\Mcal\Rcal}\Sb^{-1}_{\Rcal\Rcal} (\xc_{\Rcal} - \bm{\mu}_\Rcal), ~~~~\bm{v}_{\Mcal|\Rcal} = \Sb_{\Mcal\Mcal} + \Sb^{\top}_{\Mcal\Rcal}\Sb^{-1}_{\Rcal\Rcal} \Sb_{\Mcal\Rcal}.
\end{equation*}

\section*{C~~~Experiments}

The code for experiments is written in Python 3.9 and uses the Pytorch syntax for the automatic differentiation of the models. It can be found in the repository \url{https:github.com/pmorenoz/MPT-LML/}, where we also included the \emph{scripts} used to evaluate \textsc{bert} \citep{devlin2018bert} and the area under the \textsc{mpt} curve for different masking rates and test subsets. All figures included in the manuscript are reproducible and we also provide \emph{seeds}, the setup of learning hyperparameters as well as the initial values of parameters in the tractable model.

\subsection*{C.1~~~Longer discussion on the role of masking rates.}

The fact that fixed size \emph{held-out} sets induce a \emph{biased} estimation of the marginal likelihood in cross-validation was previously observed in \citet{pablo:neurips:2022} and \citet{fong2020marginal}. The former used this property to characterize stochastic approximations in Gaussian process models. On the other hand, the latter identified this effect as a result of the non-uniform sampling of the size of the held-out sets, e.g. setting it fixed, which in their formal results led to the biased estimate of the \emph{cumulative cross-validation} term included in the Appendix.

In this paper, we make a similar observation on the \emph{biased} estimation of the marginal likelihood in masked pre-training, where we are fixing the masking rate instead. Our empirical results with tractable models allow us to accurately identify this bias and prove that it does not affect the maximisation of the \textsc{lml}. Mainly, due to the bias is fixed during optimization (see Fig.\ 2). One important detail to consider is that this bias can be computed for tractable models, as it is the \emph{expected} log-marginal likelihood on the \emph{unmasked} tokens of the data.

\subsection*{C.2~~~Datasets}

Our experiments make use of two well-known datasets and one benchmark: \textsc{mnist} \citep{lecun1998gradient}, \textsc{fmnist} \citep{xiao2017fashion} and \textsc{glue} \citep{wang2019glue}. The datasets \textsc{mnist} and \textsc{fmnist} were downloaded from the \texttt{torchvision}
repository included in the Pytorch library. \textsc{glue} can be accessed via the public repository at \texttt{https://github.com/nyu-mll/GLUE-baselines} or \texttt{https://gluebenchmark.com/}. These particular datasets and benchmarks are not subject to use constraints related to our experiments or they include licenses which
allow their use for research purposes.

\end{document}